\documentclass[final]{cvpr}

\usepackage{times}
\usepackage{epsfig}
\usepackage{graphicx}
\usepackage{amsmath}
\usepackage{amssymb}
\usepackage{subfigure}
\usepackage{multirow}
\usepackage{enumitem}
\DeclareMathOperator*{\argmax}{arg\,max}

\usepackage[pagebackref=true,breaklinks=true,colorlinks,bookmarks=false]{hyperref}

\begin{document}

\title{Weakly-Supervised Arbitrary-Shaped Text Detection with Expectation-Maximization Algorithm}

\author{Mengbiao Zhao\textsuperscript{\rm 1,2}\quad Wei Feng\textsuperscript{\rm 1,2}\quad Fei Yin\textsuperscript{\rm 1,2}\quad Xu-Yao Zhang\textsuperscript{\rm 1,2}\quad Cheng-Lin Liu\textsuperscript{\rm 1,2,3}  \\
\textsuperscript{\rm 1} NLPR, Institute of Automation, Chinese Academy of Sciences, Beijing, P.R. China\\
\textsuperscript{\rm 2} University of Chinese Academy of Sciences, Beijing, P.R. China\\
\textsuperscript{\rm 3} CAS Center for Excellence of Brain Science and Intelligence Technology, Beijing, P.R. China\\
{\tt\small Email:\quad zhaomengbiao2017@ia.ac.cn, \{wei.feng, fyin, xyz, liucl\}@nlpr.ia.ac.cn}
}

\maketitle

\begin{abstract}
Arbitrary-shaped text detection is an important and challenging task in computer vision. Most existing methods require heavy data labeling efforts to produce polygon-level text region labels for supervised training. In order to reduce the cost in data labeling, we study weakly-supervised arbitrary-shaped text detection for combining various weak supervision forms (e.g., image-level tags, coarse, loose and tight bounding boxes), which are far easier for annotation. We propose an Expectation-Maximization (EM) based weakly-supervised learning framework to train an accurate arbitrary-shaped text detector using only a small amount of polygon-level annotated data combined with a large amount of weakly annotated data. Meanwhile, we propose a contour-based arbitrary-shaped text detector, which is suitable for incorporating weakly-supervised learning. Extensive experiments on three arbitrary-shaped text benchmarks (CTW1500, Total-Text and ICDAR-ArT) show that (1) \textbf{using only 10\% strongly annotated data and 90\% weakly annotated data, our method yields comparable performance to state-of-the-art methods,} (2) with 100\% strongly annotated data, our method outperforms existing methods on all three benchmarks. We will make the weakly annotated datasets publicly available in the future.
\end{abstract}


\section{Introduction}
Scene text detection has received increasing attention in recent years due to its wide applications in document analysis and scene understanding. Driven by the deep neural network, many previous methods have made great progress. The proposed detectors can detect not only horizontal texts~\cite{7298871, TextBoxes}, but also multi-oriented texts~\cite{DDR_2017_ICCV, Zhou_2017_CVPR} and even more challenging arbitrary-shaped texts~\cite{TextDragon_2019_ICCV, DBNet_2020_AAAI}. 

However, these methods usually require large scale polygon-level annotated data during training. According to \cite{CTW1500_2017_CVPR}, labeling one curve text consumes approximately triple time than labeling with quadrangle (13s vs 4s). Therefore, acquiring such a strongly annotated dataset is expensive and time consuming, which has impeded its application in large-scale real problems.
\begin{figure}
\centering
\includegraphics[width=3.3 in,height=2in]{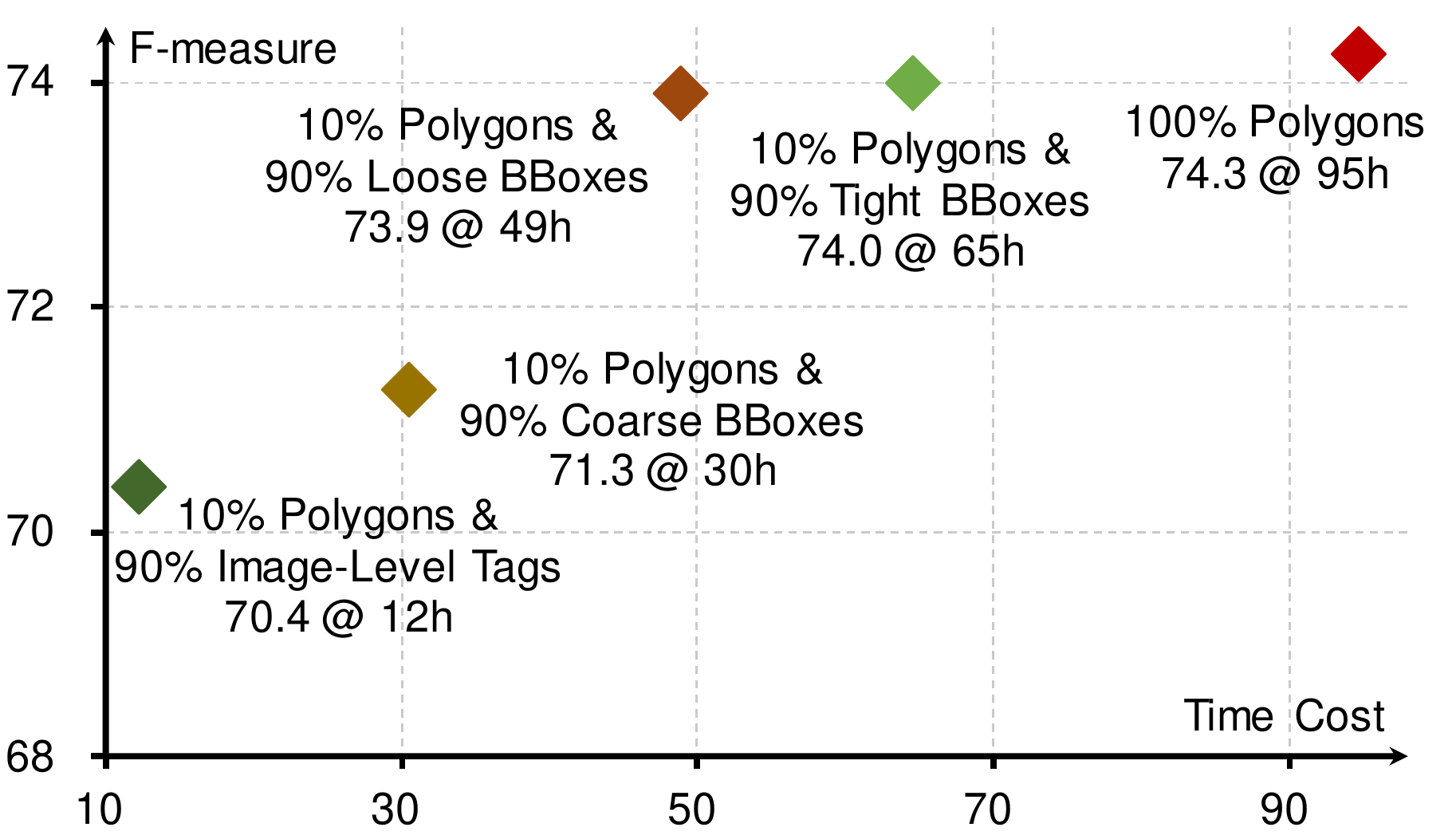}
\caption{ The comparisons of several annotation policies on the ICDAR-ArT, in terms of both accuracy (F-measure) and annotation time cost. The combination of weak annotation formats and strong annotation formats achieves the ideal tradeoff between effectiveness and time consumption.}
\label{intro}
\end{figure}
\begin{figure*}
\centering
\includegraphics[width=5.9in,height=2.55in]{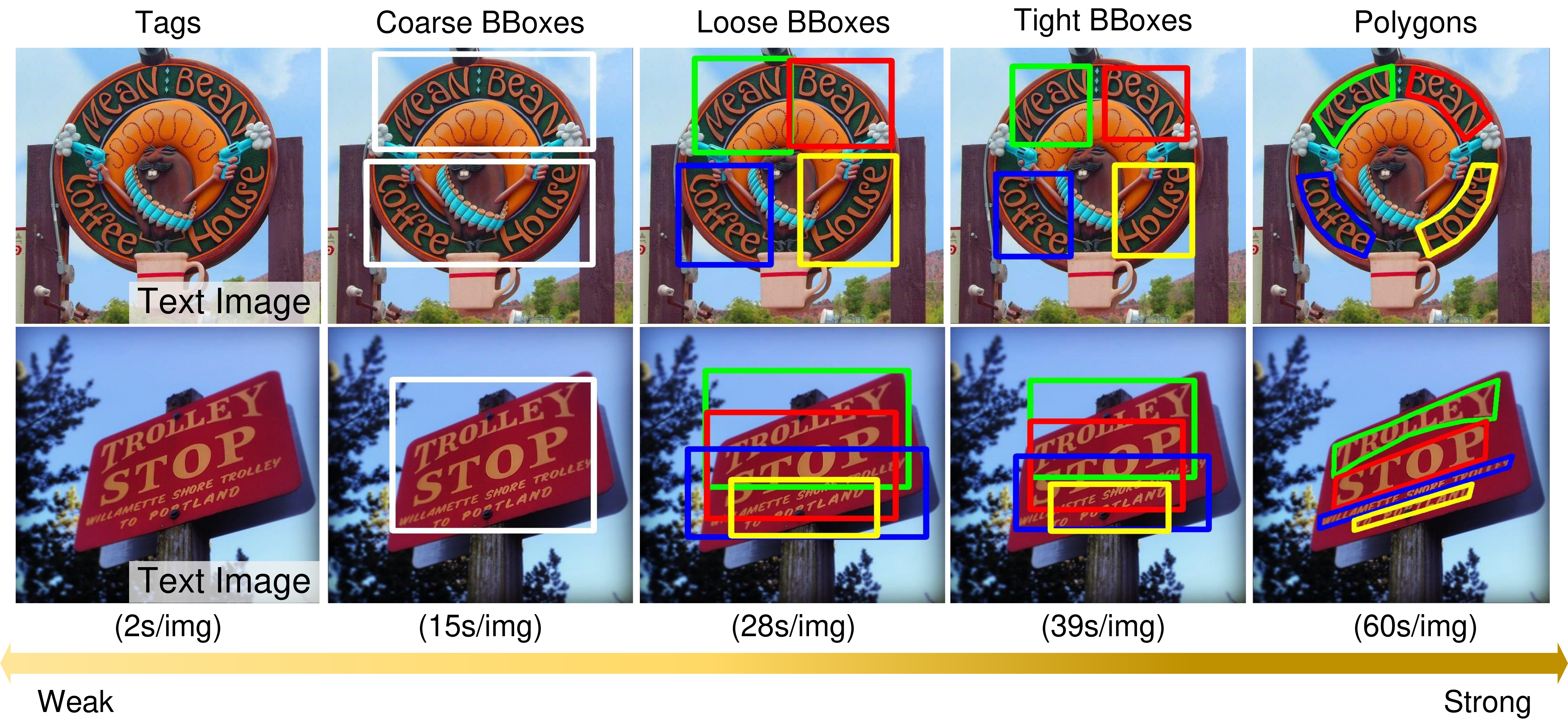}
\caption{Examples of the five supervision forms, and the time costs of labeling an image with them.
}
\label{labels}
\end{figure*}

An alternative is to utilize weak annotations. In this paper, we propose four kinds of weak supervision for arbitrary-shaped texts, which are far easier to label compared with polygons. We introduce them from strong to weak according to the supervision hierarchy. The first one is \emph{tight bounding box}, which is defined by the four extreme points of the text contour, and can enclose the text instance tightly. The second one is \emph{loose bounding box}, whose sides are not required to be close to the text instance. The third one is \emph{coarse bounding box} which locates a cluster of texts roughly. The last one is \emph{image-level tag}, which indicates whether an image contains text or not. Fig.~\ref{labels} gives an illustration of the four weak supervision forms as well as the traditional strong supervision of polygons. 

To give a comparison of time costs of labeling an image with different annotation formats, we randomly selected 500 images, and annotated them with these annotation formats by ourselves. The average time cost of each annotation format is also shown in Fig.~\ref{labels}. We can see that it takes about 1 minutes for a annotator to label an image with polygons, 39s with tight bounding boxes, 28s with loose bounding boxes, 15s with coarse bounding boxes, and only 2s with image-level tags. Based on this, we estimated the time consumptions for labeling the whole ICDAR-ArT~\cite{ArT} dataset with different annotation policies. As shown in Fig.~\ref{intro},  the time consumption can be greatly reduced by combining weak labels, which demonstrates the effectiveness and low costs of the proposed weak supervision forms. However, it still remains a challenging problem to use these weak labels to train text detectors.

In this paper, we develop an Expectation-Maximization (EM) based weakly-supervised framework for training arbitrary-shaped text detector from strong labels in combination with the proposed weak labels. Firstly, for arbitrary-shaped text detection, we propose a contour-based method, which firstly locates the instance-level text proposals, then regresses contours of them. Due to the proposal mechanism is very similar to the proposed weak supervision forms (coarse, loose, and tight bounding boxes), the detector is very suitable for incorporating weakly-supervised learning. Secondly, in order to utilize the weakly-supervised data to boost the performance, we treat polygon-level labels as latent variables for weakly-supervised data, and use a EM-like optimization algorithm (See Fig.~\ref{em}) to solve the latent variable learning problem. Specifically, the algorithm alternates between two steps: (1) E-step: estimate a probability distribution over all possible latent polygons of text instances; (2) M-step: update the weights of detection model using estimated polygon-level pseudo labels from the last E-step. In practice, the quality of the final extremum depends heavily on the initialization, since the whole optimization problem is highly non-convex. On this issue, we use the model pre-trained by a small set of strongly annotated data to initialize our learning algorithm.



In summary, this work has the following contributions: (1) We propose an EM-based framework, including a contour-based arbitrary-shaped text detector for weakly-supervised arbitrary-shaped text detection, which can accommodate various forms of weak supervision. (2) We propose four kinds of weak supervision for arbitrary-shaped text detection, which could cover most of the weakly-supervised scenarios. (3) Using only 10\% strongly annotated data combined with 90\% weakly annotated data, our model can almost match the performance of fully-supervised models, which demonstrates the superiority of our method. Meanwhile, under full supervision, our method outperforms existing methods on three public benchmarks (CTW1500~\cite{CTW1500_2017_CVPR}, Total-Text~\cite{Total-Text_2017_ICDAR}, and ICDAR-ArT~\cite{ArT}).
\begin{figure*}
\centering
\includegraphics[width=6.3in,height=1.4in]{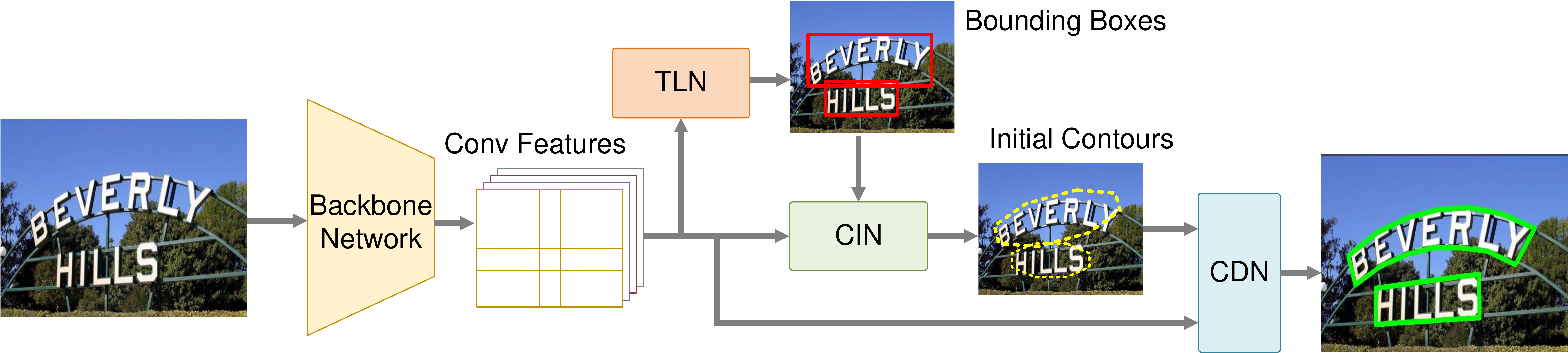}
\caption{Illustration of the structure of the proposed text detector.``TLN", ``CIN", and ``CDN" represent text localization network, contour initialization network and contour deformation network, respectively. }
\label{deepsnake}
\end{figure*}
\section{Related Work}
In this section, we review the most relevant works on scene text detection and weakly-supervised text detection.
\subsection{Scene Text Detection}
Over the years, the traditional SWT~\cite{SWT, 1315187} based methods and MSER~\cite{MSER_2010, 6248097} based methods have been replaced by deep learning inspired methods with simplified pipelines. These deep neural networks based methods \cite{2016Detecting, DDR_2017_ICCV, Zhou_2017_CVPR} are mainly built on generic object detectors \cite{FasterRCNN, SSD, DBLP:journals/corr/HuangYDY15} with various text-specific modifications.


Recently, detecting texts with arbitrary shapes has gradually drawn the attention of researchers. TextSnake~\cite{TextSnake} describes the curved text as a series of ordered, overlapping disks centered at symmetric axes. MaskTextSpotter \cite{MaskTextSpotter_2019_TPAMI} regards the arbitrary-shaped text detection as an instance segmentation problem. CRAFT \cite{CRAFT_2019_ICCV} detects individual characters, and connects them to get each text polygon. PSENet \cite{Wang_2019_CVPR} adopts a progressive scale algorithm to gradually expand the predefined kernels. LOMO \cite{Zhang_2019_CVPR} regresses text geometry in a corarse-to-fine manner. TextMountain \cite{DBLP:journals/corr/abs-1811-12786} generates text polygons after predicting text score, text center border probability, and text center direction. However, these methods usually require amounts of strongly annotated data during training. Different from these methods, we propose a novel EM-based framework, which can achieve good performance with only a small amount of fully-supervised data.

\subsection{Weakly-Supervised Scene Text Detection}
Some researchers proposed weakly-supervised learning based methods \cite{Tian_2017_ICCV, wordsup, Xing_2019_ICCV} to perform character-level text detection, due to most existing datasets do not provide character-level annotations. Usually, synthetic datasets are firstly used to initialize a character detector. Then the initial detector model could generate character candidates for the real datasets. However, such generated pseudo annotations may contain many false positives. To overcome this, the methods in \cite{wordsup} and \cite{Tian_2017_ICCV} take the polygon-level annotations as filters, and mask false positives out. Besides, the method in \cite{Xing_2019_ICCV} removes false positives by comparing the number of detected character bounding box inside the word polygon with the ground truth word. The disadvantage of these methods lies in the dependence on polygon-level annotations.

Besides, Sun \emph{et al.}~\cite{Sun_2019_ICCV} published a large partially labeled dataset (Each image is only annotated with one dominant text) and proposed an algorithm to use these partially labeled data. However, because the supervision information is too weak, it will not greatly improve the detection performance.
\section{Methodology}
The weakly-supervised framework is based on the proposed four kinds of weak supervision, which consists of two major parts: a contour-based arbitrary-shaped text detector and an EM-based learning algorithm. Firstly, we introduce the generation methods of the four weak labels for public datasets. Then, we introduce these two major parts of the framework in detail. 

\subsection{Weak Labels Generation}
Four kinds of weak labels can be generated from the polygon-level labels provided by public datasets. The \emph{tight bounding box} can be obtained from the external bounding box of the polygon. The \emph{loose bounding box} can be obtained by expanding the height and width of the tight bounding box by $0.1$ to $0.2$ times respectively. The \emph{coarse bounding box} can be generated from the external bounding box of the text cluster. We adopt Mean Shift algorithm \cite{meanshift} to cluster centers of text regions, where the clustering radius is set to 0.3 times of the short side length of the image. The \emph{image-level tag} indicates whether an image contains text or not, which can be easily obtained. 

Our purpose is to evaluate the effectiveness of the weakly-supervised text detection, therefore, the weak supervisions are generated from the polygon-level supervision provided in the public datasets. In real applications, the annotation of these weak labels are much easier compared with polygons, which will then significantly reduce the labeling costs.

\subsection{Contour-Based Text Detection}
\label{detecor}
Inspired by \cite{DeepSnake_2020_CVPR}, we propose a contour based arbitrary-shaped text detector. An overview of the detector is illustrated in Fig.~\ref{deepsnake}. After extracting original features by the backbone network, a text localization network is used to generate bounding-box-level text proposals. Then, we adopt a contour initialization network to produce the initial text contour for each text proposal. Finally, initial text contours together with original features are sent to the contour deformation network, which performs iterative contour regression to obtain the text instance boundary. Since the proposal mechanism in the pipeline is very similar to the proposed weak supervision forms (\emph{coarse}, \emph{loose}, and \emph{tight bounding boxes}), the proposed method can fully utilize the weak labels to boost the detection performance.

\vspace{1ex}
\noindent\textbf{Text Localization Network.} We adopt CenterNet \cite{DBLP:journals/corr/abs-1904-07850} to generate text proposals, which reformulates the detection task as a keypoint detection problem. The detection head has two branches: (1) The classification branch calculates a heatmap, where the peaks are supposed to be the text instance centers; (2) The regression branch predicts the height and width of the proposal bounding box for each peak. 
\begin{figure}
\centering
\includegraphics[width=3.2in,height=2.9in]{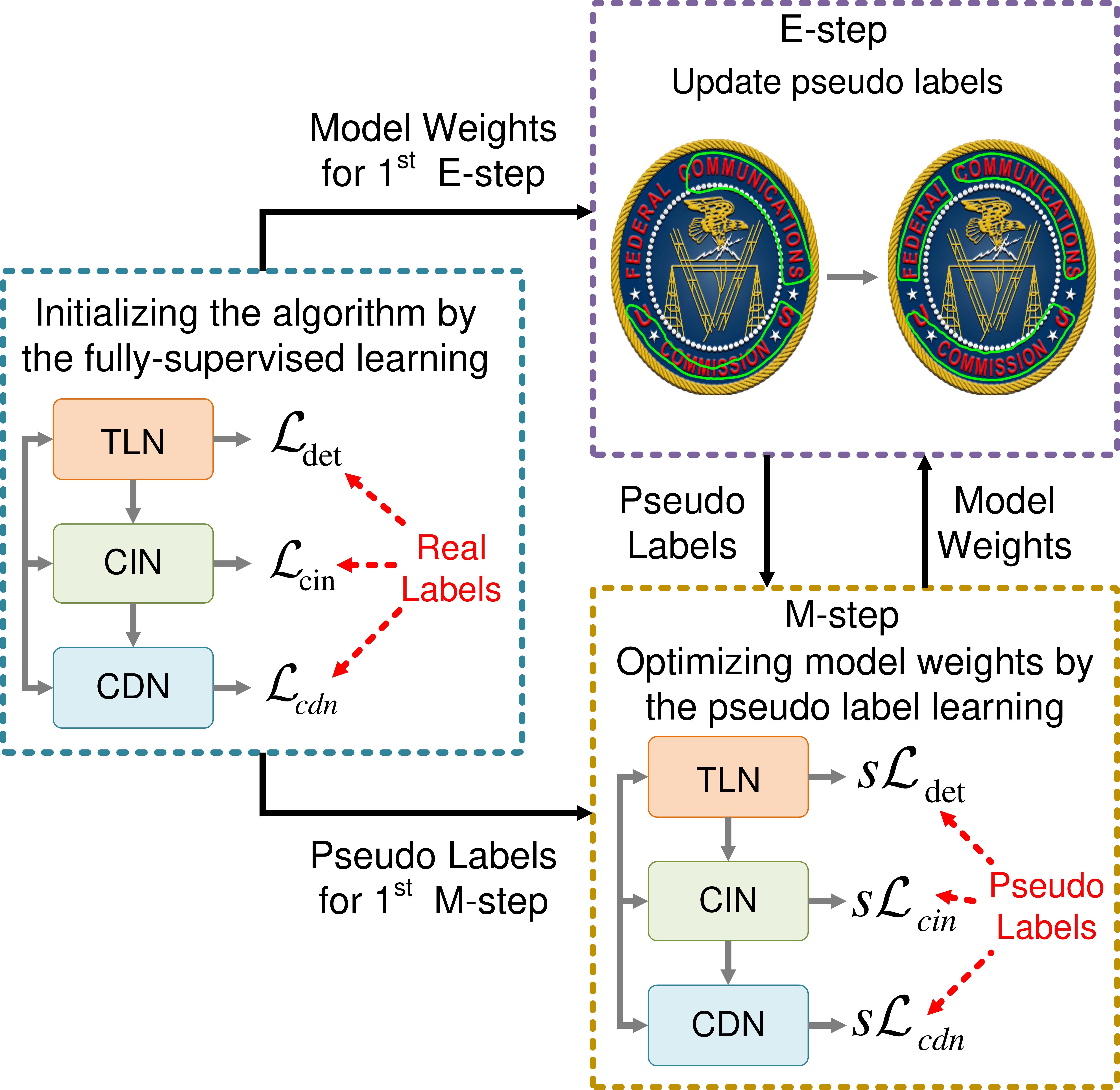}
\caption{The pipeline of the EM-based learning algorithm. The algorithm is firstly initialized by the model trained with a small amount of strongly annotated data (The blue box part). Then the algorithm alternates between updating the pseudo labels (The purple box part) and optimizing the model weights with the pseudo label learning, where the confidence weighted loss is adopted (The yellow box part).
}
\label{em}
\end{figure}

\noindent\textbf{Contour Initialization Network.} Since the initial input has a non-ignorable influence on the contour deformation and the detected text proposals usually have some offsets or errors, we propose this network to produce more accurate and suitable initial contours for text instances. In \cite{Zhou_2019_CVPR} and \cite{DeepSnake_2020_CVPR}, octagon enclose the arbitrary-shaped object much tighter than the rectangle. Therefore, we also choose it as the initial contour. In fact, the octagon could be formed by four extreme points, which are top, leftmost, bottom, rightmost pixels in an object, denoted by $\{z_{i}^{ex}|i = 1,2,3,4\}$. Therefore, the problem is how to get the extreme points from the bounding box. 

Given a bounding box, we could extract the four center points at the top, left, bottom, right box edges, denoted by $\{z_{i}^{bb}|i = 1,2,3,4\}$, and then connect them to get a diamond contour. After that, we adopt the Deep Snake, a contour regression model proposed by \cite{DeepSnake_2020_CVPR}. It takes the diamond contour as input and outputs four offsets that point from each $z_{i}^{bb}$ to the extreme point $z_{i}^{ex}$, namely $z_{i}^{ex} - z_{i}^{bb}$. Finally, we extend a line in both directions at each extreme point, whose length is 1/4 to the corresponding edge, and connect their endpoints to get the octagon. 

\noindent\textbf{Contour Deformation Network.} Contour deformation network is used to regress the offsets from points on the initial contour to the corresponding points on the ground-truth. We adopt the same regression method as the contour initialization. To make the output contour smoother, we sample 128 points along the octagon contour. Similarly, the ground-truth is generated by uniformly sampling 128 vertices along the polygon. In addition, in order to simplify the difficulty of regression, an iterative optimization strategy is adopted. Specifically, the output contour of the previous iteration is used as the initial contour of the next iteration. In our experiments, we set the number of iterations to 3.
\subsection{EM-based Learning Algorithm}
In this paper, $x$ denotes the image values, $y$ denotes the polygon-level labels of the image, and $w$ denotes the weak labels of the image. As for the weakly annotated image, we can observe the image values $x$ and the weak labels $w$, but the text instances polygons are latent variable. We have the following probabilistic graphical model:
\begin{equation}
\label{eq1}
    P(x,y,w;\theta) = P(x)P(y|x; \theta)P(w|y).
\end{equation}

Then, in order to learn the model parameters $\theta$ from the weakly annotated data, we adopt an EM-based learning strategy as follows:

\noindent{\textbf{E-step.}} The purpose of E-step is to estimate the complete-data log likelihood. Given the previously estimated parameter $\theta^{'}$, the expected complete-data log likelihood for weakly annotated image $x$ and it's label $w$ is given by
\begin{equation}
\label{eq2}
\begin{aligned}
    Q(\theta;\theta^{'}) &= \sum_{y}P(y|x,w;\theta^{'})\text{log}P(y|x;\theta) \\
    & \approx \text{log} P(\hat{y}|x;\theta),
\end{aligned}
\end{equation}
where the latent variable can be estimated by 
\begin{equation}
\label{eq3}
    \hat{y} = \argmax_{y} P(y|x,w;\theta^{'}).
\end{equation}

\noindent{\textbf{M-step.}} The M-step is to maximize the $Q(\theta;\theta^{'})$ with respect to $\theta$. According to Eq.~\ref{eq2}, the key to maximize $Q(\theta;\theta^{'})$ is maximizing $\text{log} P(\hat{y}|x;\theta)$. Here, we treat $\hat{y}$ as ground truth polygons, and optimize $\text{log} P(\hat{y}|x;\theta)$ by the mini-batch SGD algorithm.

\vspace{1 ex}We integrate the detector in Sec.~\ref{detecor} into the learning algorithm, and obtain a pipeline for weakly-supervised text detection, which is shown in Fig.~\ref{em}. The parameter $\theta$ is equivalent to the weights of the detection model. We use a small amount of strongly annotated data to train a model, which provide the initial state for the 1st M-step. And the estimated latent polygon-level label $\hat{y}$ is given by the output of contour deformation network in the detection model. As shown in Eq.~\ref{eq3}, $\hat{y}$ is related to weak labels $w$. Different weak supervisions will provide different informations, so there are different approaches to estimate the latent variables. We introduce them separately as follows:
\begin{figure}
\centering

\subfigure[]{
\begin{minipage}[t]{0.4\textwidth}
\centering

\includegraphics[width=3in,height=0.8in]{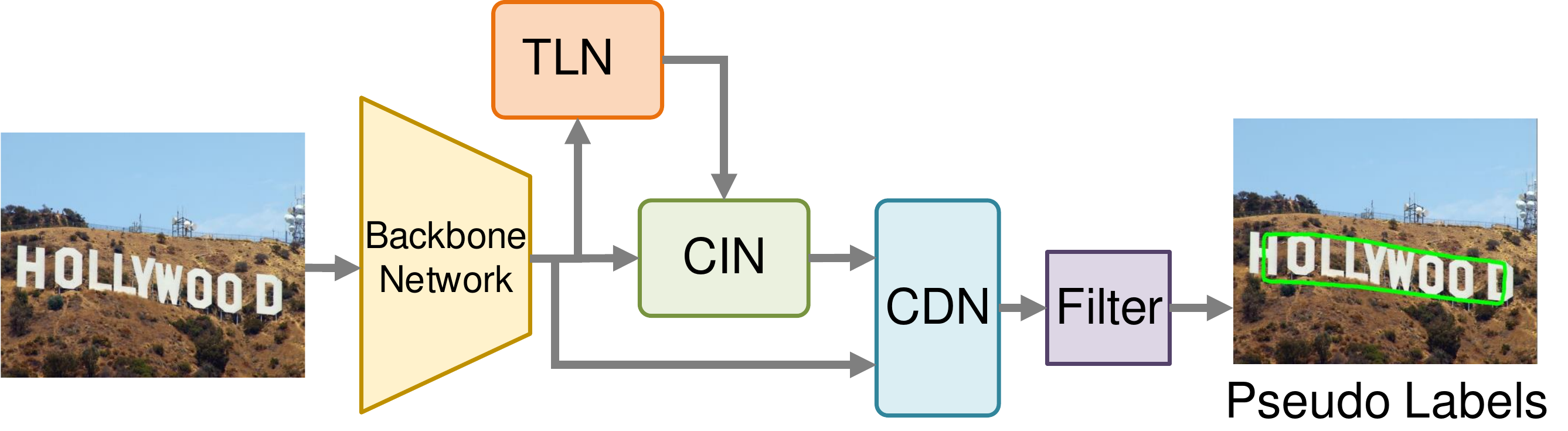}
\end{minipage}%
}%

\subfigure[]{
\begin{minipage}[t]{0.4\textwidth}
\centering

\includegraphics[width=3in,height=0.95in]{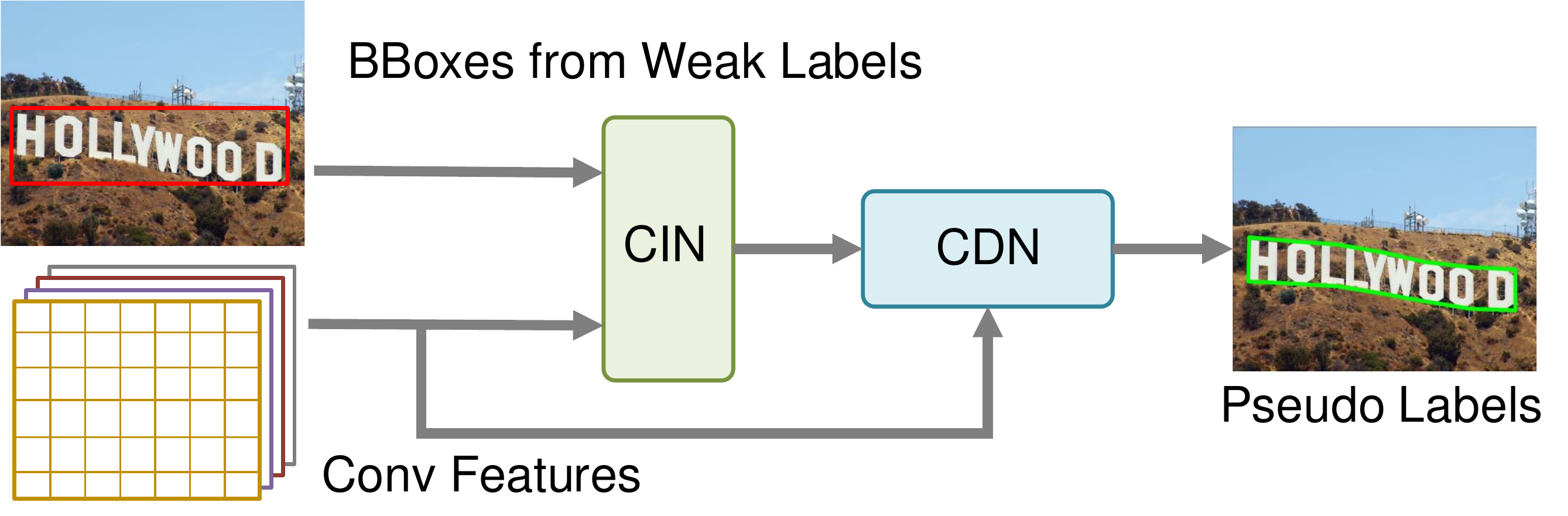}
\end{minipage}%
}%
\caption{Pseudo labels generation. (a) Pipeline for images labeled with image-level tags or coarse bounding boxes, which includes an inference step and a filtering process. (b) Pipeline for images labeled with tight or loose bounding boxes, where bounding boxes generated by text localization network are replaced with the bounding boxes from weak labels.}
\label{pesudo_gt_gen}
\end{figure}


\noindent{\textbf{Learning with Image-Level Tags.}} The whole pipeline is shown in Fig.~\ref{pesudo_gt_gen}(a). We send the text images to the detection model of the previous M-step, and obtain a candidate pseudo annotation set $D_{I}=\{(b_1,p_1,s_1),(b_2,p_2,s_2),$ $...,(b_n,p_n,s_n)\}$, where $(b_j,p_j,s_j)$ corresponds to the bounding box, polygon, and confidence score of the $j$-th detected instance, respectively. For the issue of false positives, we adopt a filter, which use the confidence threshold to select the more reliable samples. This process can be formulated as:
\begin{equation}
\label{eq4}
    D_{I}^{'} = \{(b_j,p_j,s_j)|s_j>S,(b_j,p_j,s_j)\in{D_{I}}\},
\end{equation}
where $D_{I}^{'}$ is the final pseudo annotation set, and $S$ is the confidence threshold.

\noindent{\textbf{Learning with Coarse Bounding Boxes.}} Similarly, the detection model of previous M-step is applied to the weakly annotated images (See Fig.~\ref{pesudo_gt_gen}(a)), and a candidate pseudo annotation set $D_{C}$ is obtained. In the filtering process, in addition to the confidence threshold, we can utilize the coarse bounding box which can perform like a text filter. Specifically, we use the IoU between detection results and ground truths as a metric to filter out the false positives. This process can be formulated as:
\begin{equation}
\label{eq5}
\begin{aligned}
    D_{C}^{'} = \{(b_j,p_j,s_j)|s_j>S,(b_j,p_j,s_j)\in{D_{C}}, \\
     \max_{k}\text{IoU}(b_j, g_k)>H, g_k\in{G_{C}}\},
\end{aligned}
\end{equation}
where $D_{C}^{'}$ is the final pseudo annotation set, $G_{C}$ is the ground-truth set, $S$ and $H$ is the confidence threshold and IoU threshold respectively.

\noindent{\textbf{Learning with Tight Bounding Boxes.}} Given the tight bounding boxes, the text proposal localization in the detection pipeline is unnecessary. We replace the bounding boxes generated by text localization network with the bounding boxes from weak labels (See Fig.~\ref{pesudo_gt_gen}(b)). With the ground truth bounding boxes, the more accurate initial contour are generated. Accordingly, the contour deformation network can predict better final contours. Therefore, all the generated results can be taken into final candidate pseudo annotation set $D_{T}^{'}$, and their confidence scores are all set to 1, which can be formulated as: 
\begin{equation}
\label{eq6}
    D_{T}^{'} = \{(g_k,p_k,s_k)|g_k\in{G_{T}}\},
\end{equation} 
where $D_{T}^{'}$ is the final pseudo annotation set, and $G_{T}$ is the ground truth set.

\noindent{\textbf{Learning with Loose Bounding Boxes.}} Almost the same approach as the tight bounding box is utilized in this case (See Fig.~\ref{pesudo_gt_gen}(b)). However, the loose bounding boxes contain more background noises than tight bounding boxes, which might deteriorate the performance of contour regression. Therefore, care must be taken during pre-training in this case, the generated bounding boxes of text proposals should be expanded randomly before sent to the contour initialization network, which can weaken the sensitivity of contour initialization network to the looseness of text bounding box. Similar to tight bounding boxes, all the generated results are taken into final candidate pseudo annotation set $D_{L}^{'}$ and their confidence scores are all set to 1. The final candidate pseudo annotation can be written as:
\begin{equation}
\label{eq7}
    D_{L}^{'} = \{(g_k,p_k,s_k)|g_k\in{G_{L}}\},
\end{equation}
where $D_{L}^{'}$ is the final pseudo annotation set, and $G_{L}$ is the ground truth set.

\vspace{1 ex}During network training, the same loss function as CenterNet \cite{DBLP:journals/corr/abs-1904-07850} is adopted in text localization, which can be denoted as $\mathcal{L}_{det}$. And the smooth L1 loss is used to learn contour initialization and contour deformation, which can be denoted as $\mathcal{L}_{cin}$ and $\mathcal{L}_{cdn}$, respectively. For the end-to-end training of the three tasks, the whole loss function can be formulated as:
\begin{equation}
\label{eq8}
    \mathcal{L} = \mathcal{L}_{det} + \lambda_{1}\mathcal{L}_{cin} + \lambda_{2}\mathcal{L}_{cdn},
\end{equation}
where $\lambda_{1}$ and $\lambda_{2}$ are the hyper-parameters to control the
balance among each task.

Inevitably, even if we filter the candidate pseudo annotation set generated in E-step, there are still negative samples in the final pseudo annotation set, especially for $D_{I}^{'}$ and $D_{C}^{'}$. Once using these noisy labels to update the parameters in M-step, the performance of the model tends to deteriorate iteratively. To make the training focus more on reliable samples, a confidence weighted loss $\hat{\mathcal{L}}$ is proposed as follows:
\begin{equation}
\label{eq9}
    \hat{\mathcal{L}} = s\mathcal{L},
\end{equation}
where $s$ is the confidence score of a sample obtained in the E-step, which ensures the stability of the training process.

\begin{figure*}
\centering
\includegraphics[width=5.4in,height=2.2in]{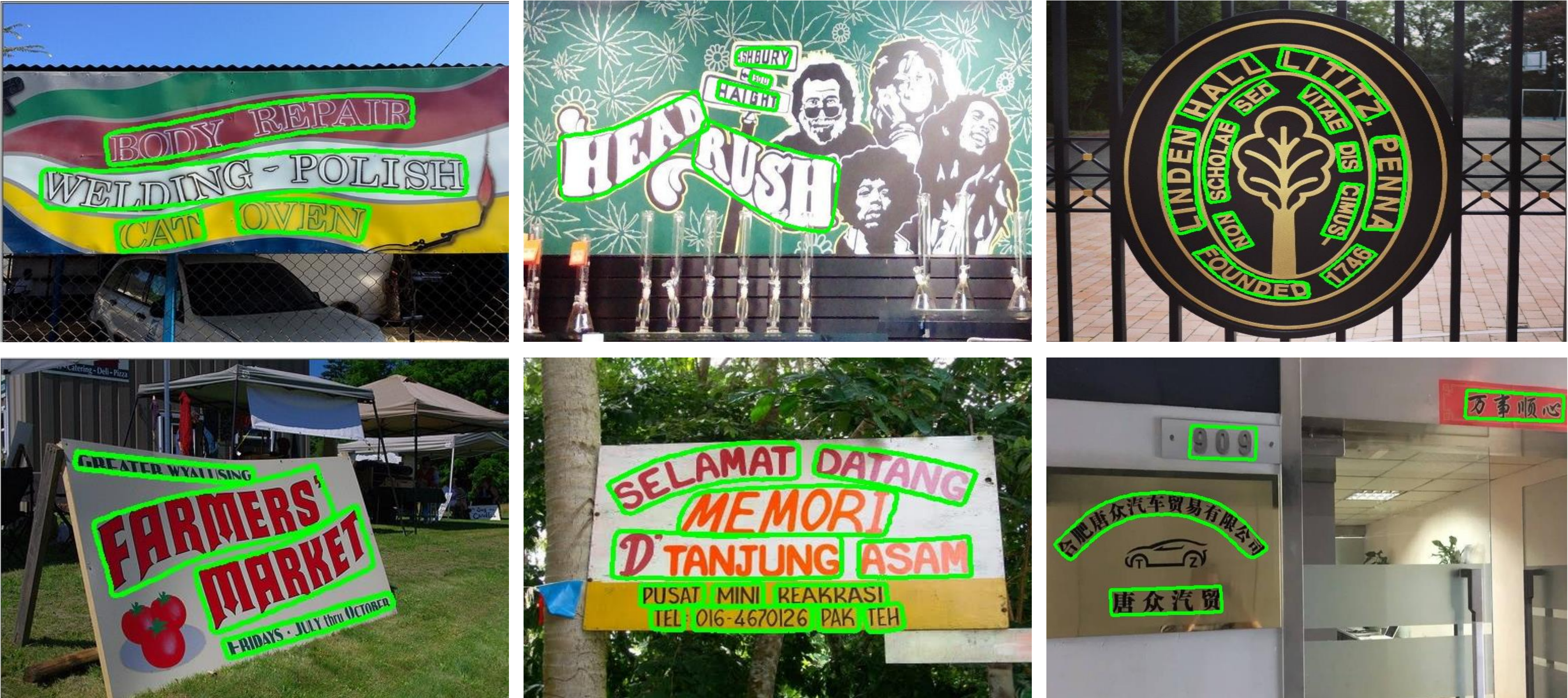}
\caption{Examples of text detection results. First column: CTW1500; second column: Total-Text; third column: ICDAR-ArT.
}
\label{res_exp}
\end{figure*}

\section{Experiments}
\subsection{Datasets}
We evaluate our method on three standard arbitrary-shaped text benchmarks: CTW1500 \cite{CTW1500_2017_CVPR}, Total-Text \cite{Total-Text_2017_ICDAR} and ICDAR-ArT \cite{ArT}.

\textbf{CTW1500.} The CTW1500 dataset contains 1000 training images and 500 test images. Each image has at least one curved text. Horizontal and multi-oriental texts are also contained in this dataset. Each text is labeled as a polygon with 14 vertexes in line-level.

\textbf{Total-Text.} The Total-Text dataset has 1255 training images and 300 test images, which contains curved texts, as well as horizontal and multi-oriental texts. Each text is labeled as a polygon with 10 vertexes in word-level.

\textbf{ICDAR-ArT.} The ICDAR-ArT dataset consits of 5603 training images and 4563 test images, which contains multi-lingual arbitrary-shaped texts. Each text is labeled with adaptive number of vertices.

\subsection{Implementation Details}
The whole algorithm is implemented with PyTorch 1.1, and we conduct all experiments on a regular workstation with 4 GeForceGRX Titan GPUs. We use the DLA-34 \cite{Yu_2018_CVPR} as our backbone. For all models, we first pre-train them with the SynthText \cite{DBLP:journals/corr/GuptaVZ16} dataset, then fine-tune the models on the corresponding real-world datasets. Each M-step consists of 200 epochs. And the batchsize is set to 36. We adopt the ``multistep" strategy to adjust the learning rate. The initial learning rate is set to $1 \times 10^{-4}$ and is divided by 2 at 80th, 120th, 150th, and 170th epoch. For data labeled with coarse bounding boxes or image-level tags, we stop training after 3 M-steps. For data labeled with tight or loose bounding boxes, since the initial pseudo annotations are of high quality, it is unnecessary to iterate multiple rounds. Actually, data augmentation is important for training, especially when there is only a small amount of strongly annotated data: (1) the image is rescaled with ratio from 0.5 to 2.0 randomly. (2) the image is horizontally flipped and rotated in range $[-10^{\circ}, 10^{\circ}]$ randomly. (3) $640\times640$ random samples are cropped from the transformed image. 

In the inference stage, the short side of the input image is scaled to a fixed length (460 for CTW1500, 640 for Total-Text, and 960 for ICDAR-ArT), with the maintained aspect ratio. 
\begin{table}    
 \centering
 \caption{Detection results on CTW1500. ``P", ``R", and ``F" represent the precision, recall and F-measure, respectively.}
 \begin{tabular}{|c|c|c|c|c|}
  \hline  
  {Method} 
  ~ & P & R & F & FPS\\
  \hline 
  {TextSnake  ~\cite{TextSnake}} & 67.9 & \textbf{85.3} & 75.6 & -\\
  \hline 
  {SegLink++ ~\cite{TANG2019106954}} & 82.8 & 79.8 & 81.3 & -\\
  \hline
  {ABCNet ~\cite{liu2020abcnet}} & 83.8 & 79.1 & 81.4 & 9.5 \\ 
  \hline
  {PSENet-1s ~\cite{Wang_2019_CVPR}} & 84.8 & 79.7  & 82.2 & 3.9 \\ 
  \hline
  {DB-ResNet-50 ~\cite{DBNet_2020_AAAI}} & 86.9  & 80.2  & 83.4 & 22.0\\ 
  \hline
  {PAN-640 ~\cite{Wang_2019_ICCV}} & 86.4 & 81.2 & 83.7 & \textbf{39.8}  \\
  \hline
  {ContourNet ~\cite{Wang_2020_CVPR}} & 83.7  & 84.1 & 83.9 & 4.5\\ 
  \hline 
  \hline 
  {100\%Poly} & \textbf{87.0} & 81.8 & \textbf{84.3} & \multirow{6}*{32.3} \\
  \cline{1-4}
  {10\%Poly \& 90\%Tight} & 86.4 & 81.1 & 83.7 & ~ \\
   \cline{1-4}
  {10\%Poly \& 90\%Loose} & 86.3 & 80.1 & 83.1 & ~ \\
   \cline{1-4}
  {10\%Poly \& 90\%Coarse} & 84.0 & 80.6 & 82.3 & ~ \\
   \cline{1-4}
  {10\%Poly \& 90\%Tag} & 83.4 & 79.1 & 81.2 & ~ \\
   \cline{1-4}
  {10\%Poly} & 81.3 & 79.1 & 80.2 & ~  \\
  \hline
 \end{tabular}  
 \label{res_ctw1500}  
\end{table}
\begin{table}    
 \centering
 \caption{Detection results on Total-Text. ``P", ``R", and ``F" represent the precision, recall and F-measure, respectively.}
 \begin{tabular}{|c|c|c|c|c|}
  \hline  
  {Method} 
  ~ & P & R & F & FPS \\
  \hline  
  {TextField ~\cite{Xu_2019}} & 81.2 & 79.9 & 80.6 & -  \\
  \hline
  {PSENet-1s ~\cite{Wang_2019_CVPR}} & 84.0 & 78.0 & 80.9 & 3.9 \\
  \hline
  {SPCNet ~\cite{Xie_Zang_Shao_Yu_Yao_Li_2019}} & 83.0 & 82.8 & 82.9 & - \\ 
  \hline
  {CRAFT ~\cite{CRAFT_2019_ICCV}} & 87.6 & 79.9  & 83.6 & - \\ 
  \hline
  {DB-ResNet-50 ~\cite{DBNet_2020_AAAI}} & 87.1  & 82.5  & 84.7 & 32.0\\ 
  \hline
  {PAN-640 ~\cite{Wang_2019_ICCV}} & \textbf{89.3} & 81.1 & 85.0 & \textbf{39.6}  \\
  \hline
  {ContourNet ~\cite{Wang_2020_CVPR}} & 86.9  & \textbf{83.9}  & 85.4 & 3.8 \\ 
  \hline 
  \hline 
  {100\%Poly} & 88.2 & 83.3 & \textbf{85.6} & \multirow{6}*{24.2} \\
  \cline{1-4}
  {10\%Poly \& 90\%Tight} & 85.4 & 83.8 & 84.6 & ~  \\
  \cline{1-4}
  {10\%Poly \& 90\%Loose} & 86.6 & 82.1 & 84.3 & ~ \\
  \cline{1-4}
  {10\%Poly \& 90\%Coarse} & 84.7 & 79.6 & 82.0 & ~ \\
  \cline{1-4} 
  {10\%Poly \& 90\%Tag} & 82.9 & 78.8 & 80.8 & ~  \\
  \cline{1-4}
  {10\%Poly} & 80.2 & 78.5 & 79.4 & ~  \\
  \hline 
 \end{tabular}  
 \label{res_total-text}  
\end{table}
\begin{table}    
 \centering
 \caption{Detection results on ICDAR-ArT. ``P", ``R", and ``F" represent the precision, recall and F-measure, respectively.}.  
 \begin{tabular}{|c|c|c|c|c|}   
  \hline  
  {Method} 
  ~ & P & R & F & FPS \\
  \hline  
  {TextRay ~\cite{10.1145/3394171.3413819}} & 76.0 & 58.6 & 66.2 & -\\
  \hline 
  \hline 
  {100\%Poly} & 80.8 & \textbf{68.7} & \textbf{74.3} & \multirow{6}*{\textbf{16.4}}  \\
  \cline{1-4}
  {10\%Poly \& 90\%Tight} & 81.9 & 67.5 & 74.0 & ~ \\
  \cline{1-4}
  {10\%Poly \& 90\%Loose} & \textbf{82.0} & 67.3 & 73.9 & ~  \\
  \cline{1-4}
  {10\%Poly \& 90\%Coarse} & 77.7 & 65.8 & 71.3 & ~  \\
  \cline{1-4} 
  {10\%Poly \& 90\%Tag} & 77.9 & 64.2 & 70.4 & ~  \\
  \cline{1-4}
  {10\%Poly} & 77.2 & 58.9 & 66.8 & ~  \\

  \hline 
 \end{tabular}  
 \label{res_art}  
\end{table}

\subsection{Experimental Results}
For all three datasets, we randomly select 10\% of original training images as strongly annotated data and take other 90\% as weakly annotated data, resulting in a 100 and 900 split for CTW1500, a 125 and 1130 split for Total-Text, and a 560 and 5043 split for ICDAR-ArT. Based on the data division, we can get the following models:

(1) \emph{100\%Poly}: Model trained with all images, which are annotated with polygons.

(2)  \emph{10\%Poly}: Model trained with 10\% images, which are annotated with polygons.

(3)  \emph{10\%Poly \& 90\%XXX}: Model trained with all images, of which 10\% are annotated with polygons, and 90\% are annotated with a kind of weak annotation format.




\begin{table}    
 \centering
 \caption{The benefits of the confidence weighted loss on ICDAR-ArT. ``WL" indicates confidence weighted loss. ``P", ``R", and ``F" represent the precision, recall and F-measure, respectively.}  
 \begin{tabular}{|c|c|c|c|c|}   
  \hline  
  {Method} 
  ~ & WL & P & R & F  \\
  \hline  
  {10\%Poly \& 90\%Coarse} &-& \textbf{80.4} & 60.8 & 69.2 \\
  \hline 
  {10\%Poly \& 90\%Coarse} &$\checkmark$ & 77.7 & \textbf{65.8} & \textbf{71.3}  \\
  \hline 
  \hline
  {10\%Poly \& 90\%Tag} &-& 76.5 & 61.0 & 67.9 \\
  \hline 
  {10\%Poly \& 90\%Tag} &$\checkmark$ & \textbf{78.0} & \textbf{64.2} & \textbf{70.4}  \\
  \hline
 \end{tabular}  
 \label{weighted}  
\end{table}
\vspace{1 ex}
\noindent\textbf{Weakly-Supervised Detection Results.} As shown in Tables~\ref{res_ctw1500}, \ref{res_total-text}, and \ref{res_art}, the ``10\%Poly \& 90\%Tight" has gained the best performance among the four weakly-supervised models, and almost reaches the performance of the ``100\%Poly". Due to the loose bounding boxes contains more background noises, the performance of  ``10\%Poly \& 90\%Loose" is slightly lower than that of ``10\%Poly \& 90\%Tight". Both ``10\%Poly \& 90\%Coarse"  and ``10\%Poly \& 90\%Tag" outperform the baseline model ``10\%Poly", which demonstrates the effectiveness of the proposed weakly-supervised framework.

\vspace{1 ex}
\noindent\textbf{Fully-Supervised Detection Results.} As shown in Tables~\ref{res_ctw1500},~\ref{res_total-text}~and~\ref{res_art}, our fully-supervised models have achieved state-of-the-art performance on all three datasets, which demonstrates the superiority of our method. Meanwhile, in terms of speed, our method also achieves good performance. Although not the highest, our method achieves an impressive trade-off between speed and accuracy. Some examples of text detection results are shown in Fig.~\ref{res_exp}, where it is seen that the detection model can accurately locate horizontal, multi-oriented and curved texts.

\begin{table*}    
 \centering
 \caption{Detection results of different annotation policies on ICDAR-ArT. ``P", ``R", and ``F" represent the precision, recall and F-measure, respectively. In the ``Labels" column, the ``I", ``C", ``L", ``T" and ``Poly" represent the image-level tags, coarse bounding boxes, loose bounding boxes, tight bounding boxes and polygons, respectively.}  
 \begin{tabular}{|c|c|c|c|c|c|}   
  \hline  
  {Policy} 
  ~  & Image amount & Labels & P & R & F  \\
  \hline  
  {Strong}  & 710 & Poly & 79.5 & 59.7 & 68.2 \\
  \hline 
  {Equal Time}& 560+58+81+152+1143 & Poly+T+L+C+I & \textbf{80.3} & \textbf{65.3} & \textbf{72.1} \\
  \hline
  {Equal Number} & 560+108$\times$4 & Poly+T+L+C+I & 78.7 & 63.7 & 70.4 \\
  \hline 
  \hline 
  \multirow{4}*{ 80\% Poly} &  568+219 & Poly+T & 77.1 & 63.8 & 69.8 \\
  \cline{2-6}
   ~&  568+307 & Poly+L & 78.2 & 64.1 & 70.5 \\
  \cline{2-6}
   ~&  568+576 & Poly+C & 77.1 & 63.2 & 69.5 \\
  \cline{2-6}
   ~&  568+4320 & Poly+I & 75.8 & 64.4 & 69.6 \\
  \hline
 \end{tabular}  
 \label{budget}  
\end{table*}


\begin{figure}
\centering
\includegraphics[width=2.9in,height=1.7in]{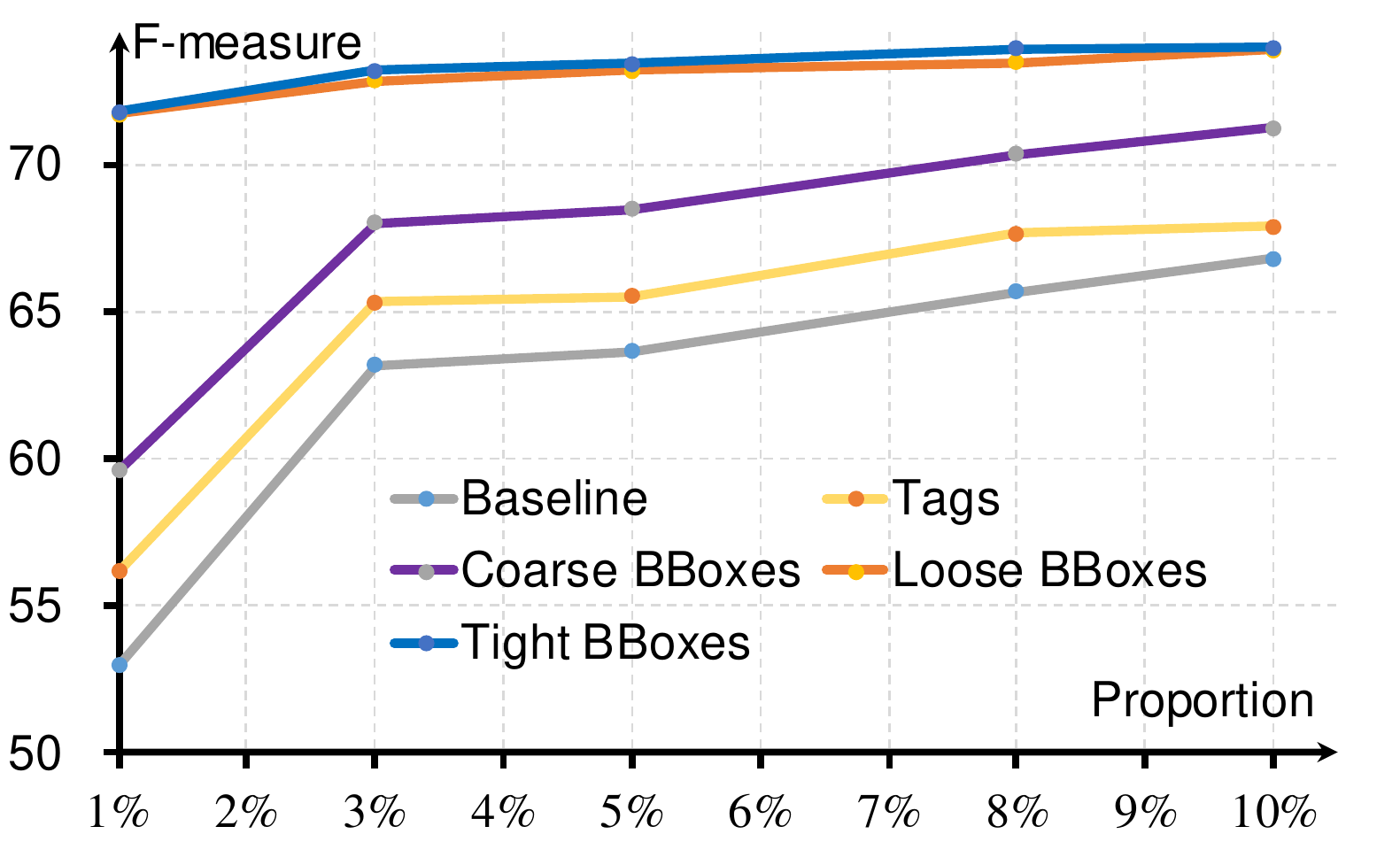}
\caption{ F-measure versus proportions of strongly annotated data for four weakly-supervised models on ICDAR-ArT test set. The baseline model is trained with only a certain proportion of strongly annotated data. The other four weakly-supervised models are trained with a certain proportion of strongly annotated data and the remaining weakly annotated data (The legends represent the weak annotation formats).
}
\label{diff}
\end{figure}
\subsection{Ablation Studies}

\noindent\textbf{Influence of the Proportion of Strongly Annotated Data.} As illustrated above, the weakly-supervised model can almost match the performance of the fully-supervised model, using only 10\% strongly annotated data. With less strongly annotated data, can the proposed method achieve such good results? We sample five proportions from 1\% to 10\% for the strongly annotated data, which are 1\%, 3\%, 5\%, 8\%, and 10\%, and conduct experiments on ICDAR-ArT. As shown in Fig.~\ref{diff}, the performance gets better as the proportion of strongly annotated data increases. In addition, the four kinds of weak supervision show different sensitiveness about the amount of strongly annotated data. For tight or loose bounding boxes annotated data, combining with only 1\% strongly annotated data, they can still get good performance. While, for the other two weak supervision forms, when the proportion of strongly annotated data is reduced, the performance drops obviously.  

\noindent\textbf{Influence of the Confidence Weighted Loss.} We evaluate the influence of the weighted confidence loss (See Eq.~\ref{eq9}) by experiments with or without it. We conduct two groups of comparative experiments on ICDAR-ArT. As shown in Table~\ref{weighted}, the models trained with confidence weighted loss yield better results. The confidence weighted loss improves the two basic models by about 2.1\% and 2.5\% of F-measure respectively. It attributes to that with confidence weighting, the influence of noise samples on training can be limited, and the network learning tends to focus on reliable samples.

\begin{figure}
\centering
\includegraphics[width=2.9in,height=1.7in]{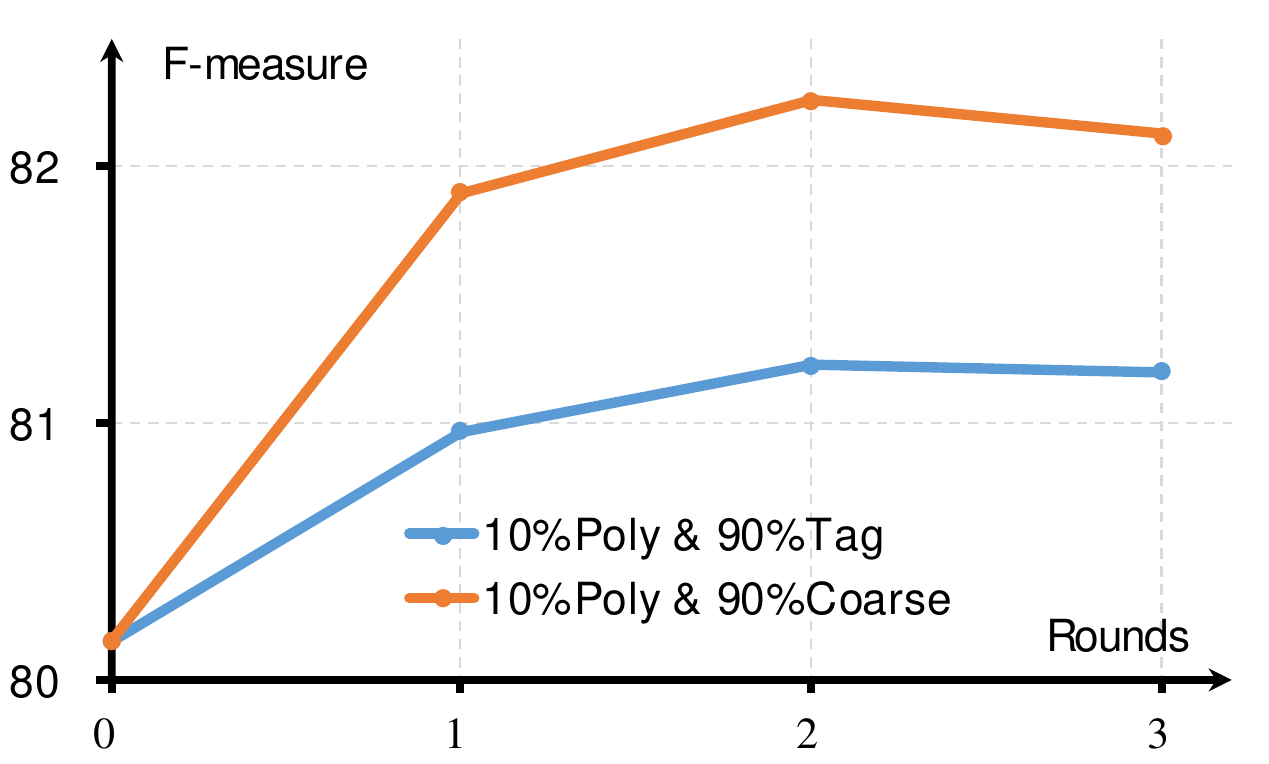}
\caption{ F-measure versus training rounds for two weakly-supervised models on CTW1500 test set.
}
\label{rounds}
\end{figure}

\vspace{1 ex}
\noindent\textbf{Influence of the Number of Training Rounds.} The results of the weakly-supervised training within three rounds are shown in Fig.~\ref{rounds}. The result of the 0 training round is the performance of the initial model. The performance of weakly-supervised models has improved in the first two rounds, which attributes to the improvement of pseudo label quality. Then the improvements stop at the third round, which lies in the error accumulation. 

\vspace{1 ex}
\noindent\textbf{Budget-Aware Omni-Supervised Text Detection.} In this part, we consider another scenario: given the annotation budget, how to design an annotation policy for better detection performance. For data annotation, the budget consumption is directly proportional to the time cost. Therefore, we just use the time cost as the budget consumption and ignore other factors.

The total budget is set to 43,200 seconds (12 hours). Then, we put forward three policies as list below: (1) \textbf{Strong}: all the budget is used to annotate polygons. (2) \textbf{Equal Time}: in addition to 560 strongly labeled images, the remaining time is equally divided into four weak supervision forms. (3) \textbf{Equal Number}: in addition to 560 strongly labeled images, the remaining time is used for the weak supervision forms, and make the image amount of each kind of weak supervision the same. In addition, in order to explore the cost-effectiveness of the four weak supervision forms, we conduct four comparative experiments, in which 80\% budget is spent on polygons and the rest is spent on a kind of weak supervision. 

The results are shown in Table~\ref{budget}, which show that both the ``Equal Time" and ``Equal Number" policy perform better than the ``Strong" policy, and the ``Equal Time" policy has gained the best results, outperforming the ``Strong" policy over 3.9\% on F-measure. These results suggest that spending a certain amount of budget to annotate more images with weak labels is better than commonly all adopted strong labels. For the ``80\% Poly" policy, the four weak supervision forms have achieved similar results, and ``Loose Bounding Box" is slightly ahead of others, which can be regarded as the most cost-effective weak supervision form.

\section{Conclusion}
In this paper, we propose an EM-based framework for weakly-supervised arbitrary-shaped text detection, so as to take advantage of various forms of weak annotations in addition to polygon-level strong annotation. The proposed framework consists of a contour-based arbitrary-shaped text detector and an EM-based learning strategy. Extensive experiments have been conducted to demonstrate the effectiveness of the proposed framework. Using 10\% strongly annotated data, our weakly-supervised model can match the performance of fully-supervised models. Meanwhile, under full supervision, our method outperforms existing methods. The results also reveal the effects of different kinds of weak supervision. How to better exploit weakly annotated data deserves further research.

{\small
\bibliographystyle{ieee_fullname}
\bibliography{references}
}

\end{document}